\documentclass{article}

\usepackage{microtype}
\usepackage{graphicx}
\usepackage{subfigure}
\usepackage{booktabs} 

\usepackage{hyperref}

\usepackage[accepted]{icml2024}

\usepackage[frozencache=true,cachedir=minted-cache]{minted}
\definecolor{LightGray}{gray}{0.9}

\usepackage{adjustbox}
\usepackage{multirow}

\renewcommand{\footnotesize}{\fontsize{8.8}{8.6}\selectfont} 

\definecolor{clightgray1}{rgb}{0.9,0.9,0.9} 
\definecolor{clightgray2}{rgb}{0.85,0.85,0.85}
\definecolor{clightgray3}{rgb}{0.8,0.8,0.8}
\definecolor{clightgray4}{rgb}{0.75,0.75,0.75}

\usepackage{amsmath}
\usepackage{amssymb}
\usepackage{mathtools}
\usepackage{amsthm}
\usepackage{bm}

\usepackage[capitalize,noabbrev]{cleveref}

\theoremstyle{plain}

\theoremstyle{definition}

\theoremstyle{remark}

\usepackage[textsize=tiny]{todonotes}

\icmltitlerunning{GISTEmbed: Guided In-sample Selection of Training Negatives for Text Embedding Fine-tuning}

\begin{document}

\twocolumn[
\icmltitle{
GISTEmbed: Guided In-sample Selection of Training Negatives \\ for Text Embedding Fine-tuning}



\icmlsetsymbol{equal}{*}

\begin{icmlauthorlist}
\icmlauthor{Aivin V. Solatorio}{yyy}
\end{icmlauthorlist}

\icmlaffiliation{yyy}{The World Bank, Office of the Chief Statistician}

\icmlcorrespondingauthor{Aivin V. Solatorio}{asolatorio@worldbank.org}

\icmlkeywords{GISTEmbed, In-sample negatives mining, Text Embeddings, Machine Learning, Deep Learning, Transformers, Sentence Transformers, ICML}

\vskip 0.3in
]



\printAffiliationsAndNotice{}  

\begin{abstract}


Embedding models are integral to AI applications like semantic search, personalized recommendations, and retrieval augmented generation for LLMs, necessitating high-quality training data. However, the limited scalability of manual data curation prompts the need for automated methods to ensure data integrity. Traditional unsupervised triplet mining automates training data generation, crucial for embedding model training, yet inadvertently injects biases and noise, thereby degrading model performance. Addressing this, we introduce \textbf{\texttt{GISTEmbed}}, a novel strategy that enhances in-batch negative selection during contrastive training through a guide model. This approach departs from reliance on random sampling and equal utility assumption of batch negatives, significantly reducing noise from data quality issues and improving model fine-tuning. Benchmarked against the Massive Text Embedding Benchmark (MTEB), GISTEmbed showcases consistent performance improvements across various model sizes and achieves state-of-the-art results in select categories. This framework enables significant enhancements for smaller models by leveraging the capabilities of powerful yet resource-intensive large models. GISTEmbed can potentially revolutionize the creation of highly efficient, smaller models, democratizing access to advanced AI technologies. Making these technologies more accessible and cost-effective, especially for applications constrained by resources, significantly expands the impact and accessibility of state-of-the-art AI solutions across diverse sectors.



\end{abstract}

\begin{figure}[ht!]
    \begin{center}
    \centerline{\includegraphics{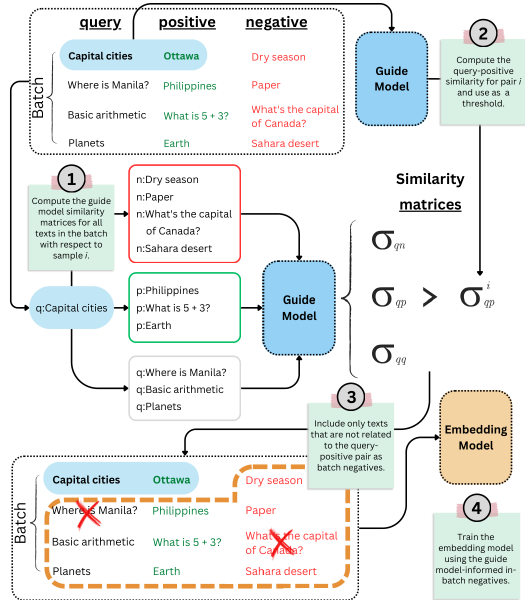}}
    \caption{Visualization of the GISTEmbed framework for the dynamic selection of in-batch negatives for contrastive learning of embedding models. A guide model is used during training to dynamically exclude texts in the batch that are likely related to the query-positive pair being evaluated. The framework addresses potential data labeling issues and also relaxes assumptions regarding the formation of in-batch negatives that prior approaches use.
    }
    \label{fig:gistembed}
    \end{center}
    \vskip -0.3in
\end{figure}

\begin{figure*}[ht!]
    \vskip 0.2in
    \begin{center}
    \centerline{\includegraphics{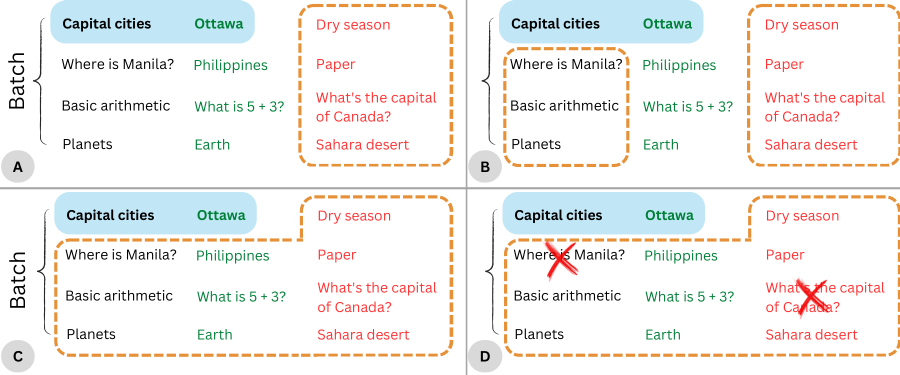}}
    \caption{Visualization of various in-batch negatives selection strategies for contrastive learning (dashed orange boxes). Each panel contains triplets in a training batch, with the columns representing the queries, assigned positives, and assigned negatives. Panel A shows the original strategy for selecting in-batch negatives where all the assigned negatives in the training data are considered. Panel B visualizes the selection of in-batch negatives for the bi-directional InfoNCE loss which includes the queries as well. The full-sample selection of in-batch negatives is shown in Panel C. While Panel D presents how GISTEmbed, with the guide model-informed selection of in-batch negatives, works. In this example, the query-positive pair \texttt{(q:Capital cities, p:Ottawa)} can be considered semantically related to the other texts in the batch \texttt{[q:"Where is Manila?", n:"What is the capital of Canada?"]}. The guide model serves as a filter to remove these texts when selecting the in-batch negatives for computing the loss.}
    \label{fig:multi-negatives}
    \end{center}
    \vskip -0.2in
\end{figure*}

\section{Introduction}
\label{introduction}


Text embedding models are essential for natural language processing (NLP), acting as critical components for a wide array of artificial intelligence (AI) applications \cite{NIPS2015_f442d33f, hill-etal-2016-learning}. These models are key to various use cases ranging from semantic search \cite{muennighoff2022sgpt}, which helps improve the accuracy of search engines, to personalized and content recommendations \cite{okura2017embedding} and enhancing the functionality of large language models (LLMs) through retrieval augmented generation (RAG) \cite{lewis2020retrieval}. By converting textual data into a format that machines can easily interpret, ensuring that meaning is encoded, text embedding models unlock the utility of abundant textual data resources for practical applications. Consequently, the data quality used in training these models is paramount, as it directly affects model performance and downstream applications. Thus, securing high-quality training data is vital for creating robust and reliable models.

However, as embedding models become increasingly central to AI systems, the scalability of data curation to produce high-quality data emerges as a non-trivial challenge. Manual data curation, while effective, is impractical for meeting the exponential growth in data requirements posed by advanced models. This gap underscores the necessity for automated methods to detect and mitigate data quality issues. Additionally, the conventional reliance on unsupervised triplet mining for generating training data introduces its own set of challenges, such as methodological biases and noise \cite{wu-etal-2022-esimcse}, which can degrade model performance.

State-of-the-art approaches in embedding model training have sought to address these challenges through various means, yet gaps remain. The common practice of indiscriminately using in-batch negatives in contrastive training, for instance, fails to account for the varied utility of these negatives, often leading to the introduction of noise and biases in the training process. Attempts at improving strategies for generating training data with LLMs have shown promising results \cite{cheng2023improving}. Despite advancements, the need for a more refined strategy to address potential data quality issues without compromising scalability or efficiency remains unmet, highlighting a critical area for innovation.

This paper introduces GISTEmbed, a novel strategy designed to improve the selection of in-batch negatives in contrastive training. Integrating a guide model to select negative samples during the training dynamically, GISTEmbed significantly reduces the reliance on random sampling and the flawed assumption that all batch negatives possess equal utility \cite{zhou-etal-2022-debiased}. This approach mitigates noise introduced by data quality issues and enhances the fine-tuning process, leading to the development of state-of-the-art models across various sizes. Our comprehensive evaluation, benchmarked against the MTEB \cite{muennighoff2022mteb}, showcases GISTEmbed's ability to consistently improve model performance, establishing a new framework that leverages the capabilities of large, high-performing guide models to augment the training efficiency and effectiveness of smaller models. Through GISTEmbed, we highlight the importance of high-quality data and the opportunities to leverage existing models to address data quality issues. This provides an alternative view in embedding model training and setting new benchmarks in the field.

\section{Overview and Motivation}

This section provides a brief overview of methods for training embedding models. Then, we will highlight implicit assumptions regarding data quality currently assumed by existing methods. We will subsequently identify potential issues arising from these assumptions, which inadvertently lead to the development of suboptimal embedding models.


\subsection{Training embedding models}

Encoding texts into vector representations has a long history. Arguably, one of the most important breakthroughs in embedding models is the invention of Word2Vec \cite{mikolov2013efficient}. It hinted at the possibility of using large-scale text data to learn meaningful semantic representations of texts unsupervisedly.

Contemporary methods for generating embedding models typically employ transformer models \cite{reimers2019sentence,INSTRUCTOR}. An encoder model is used to encode the text into vectors. The embedding vectors are usually derived by mean-pooling the token representations at the last hidden state of a transformer model \cite{ma2019universal}. Some also use only the final representation for the \texttt{[CLS]} token as the embedding for the entire text \cite{huang2021whiteningbert}. While pre-trained models have been shown to provide reasonable representations of the semantic meaning of texts using these simple pooling strategies, downstream fine-tuning provides significant improvement \cite{reimers2019sentence}.

One of the most common methods of training models for embeddings is through unsupervised contrastive learning \cite{yan-etal-2021-consert, zhang2023retrieve, xu2023contrastive}. A training architecture is leveraged, typically variants of siamese network architectures, to fine-tune embedding models \cite{reimers2019sentence}. Fine-tuning models require training data. The training data is constructed comprising pairs of texts resembling some notion of similarity or relevance. Previous studies show that having negative examples helps in generating robust text representations \cite{cheng2023improving}. Consequently, triplet mining---a method to generate triplets of text comprising a query (anchor), a positive example, and a negative example---has become an essential component in learning effective embeddings. Having negative examples makes contrastive learning possible, which helps models differentiate between relevant and irrelevant texts.

In the subsequent section, we delve into the mechanisms and strategies that enable the model to learn meaningful representations. Furthermore, we examine critical assumptions and identify potential challenges in selecting negative examples.

\subsection{InfoNCE loss and multiple-negatives}

Loss functions are instrumental in embedding desired properties into models. The prevalent approach for training embedding models often employs the InfoNCE loss \cite{oord2018representation}, which is fundamentally aligned with the contrastive loss function that incorporates multiple negatives, as introduced by \citeauthor{henderson2017efficient} for response suggestion applications. This particular loss function enables the utilization of in-batch samples for negative sampling, thereby facilitating contrastive training. The mathematical representation of the InfoNCE loss is given by:

\begin{equation}
\label{eq:infonce}
\mathcal{L} \sim \frac{e^{\text{sim}(q_i,p_i^+) / \tau}}
{ e^{\text{sim}(q_i,p_i^+) / \tau} + \sum_{j \in B} e^{\text{sim}(q_i,p_j^-) / \tau}}.
\end{equation}

This formula promotes the model to distinguish between related pairs ($q_i$, $p_i^+$) and unrelated samples $p_j^-$ within the embedding space, as quantified by a similarity metric \texttt{sim}. Cosine similarity is often the metric for measuring proximity in text embedding scenarios, while $B$ is the universe of batch negatives from which $p_j^-$ are drawn. The parameter $\tau$, known as the temperature, modulates the concentration of the similarity scores. Typically, $\tau$ values are selected within the range of [0.01, 0.1], optimizing the model's performance by adjusting the scale of similarity distributions.

Variants of negative sampling strategies to improve the loss include bidirectional contrastive loss \cite{ni2021large, INSTRUCTOR}, which also considers other queries in the batch as negatives for the given query-positive pair. Another strategy proposes using the full sample in a batch, not associated with the query or positive, as the multiple negatives \cite{karpukhin2020dense,xiao2023c,li2023towards}. An analysis of the impact of multiple negatives in contrastive learning has recently been explored by \citeauthor{cao2022exploring}. A visualization of the various in-batch negative sampling strategies is illustrated in Figure \ref{fig:multi-negatives}.

\subsection{Potential issues}

While models trained on methods we discussed above showed state-of-the-art performance, it is easy to see some potential issues concerning negative sampling in the above strategies. Mainly, it is possible that some randomly selected text in another triplet within the same batch may be relevant to the query-positive pair being evaluated. If we use such examples as negatives, the model will likely learn a suboptimal representation of the embeddings. Also, some data may be incorrect for various reasons; for example, a query may be more similar to the assigned negative in the training data than the assigned positive. We find examples of this in the raw MEDI dataset \cite{INSTRUCTOR}, Annex \ref{ax:invalidmedi}.

Ultimately, these problems can be mitigated by producing highly curated training data. However, the scale and resources needed for curation may be lacking in most settings. We look at intelligent agents or guide models as a proxy for manual data curation, ensuring scalability.

In the next section, we present a framework we call GISTEmbed---Guided In-sample Selection of Training Negatives---for training embedding models, which aims to help address the issues above.

\section{GISTEmbed Framework}

We propose GISTEmbed as a framework that leverages a guide model ($G$) to dynamically select the in-sample negatives for the contrastive training of embedding models. The guide model can be some arbitrary model or agent capable of scoring the relevance of some text relative to a given query or another piece of text. In our case, we use a large and high-performing embedding model as the guide for fine-tuning smaller embedding models--- \href{https://huggingface.co/WhereIsAI/UAE-Large-V1}{WhereIsAI/UAE-Large-V1}, a model trained using the angle-optimized (AnglE) loss function \cite{li2023angle}\footnote{\href{https://huggingface.co/WhereIsAI/UAE-Large-V1}{https://huggingface.co/WhereIsAI/UAE-Large-V1}}.

The loss function remains in form as prescribed in Equation \ref{eq:infonce}, but GISTEmbed proposes a different universe of negative examples. The modified loss function becomes:

\begin{equation}
\label{eq:lg}
\mathcal{L}_G \sim \frac{e^{\text{sim}(q_i,p_i^+) / \tau}}
{ e^{\text{sim}(q_i,p_i^+) / \tau} + \sum_{j \in G_B} e^{\text{sim}(q_i,p_j^-) / \tau}}.
\end{equation}

The GISTEmbed loss ($\mathcal{L}_G$) adopts $G_B$ as the guide model-informed batch negatives in this framework. In the next section, we discuss the process of generating $G_B$.


\subsection{Proposed strategy}

During training, the model ($M$) receives a batch of text containing the queries, their corresponding positives, and the assigned negatives. We compute the vectors for texts in this batch and compute the following pairwise cosine similarity matrices: query-positive similarities ($S_{qp}$), query-negative similarities ($S_{qn}$), query-query similarities ($S_{qq}$), and the positive-positive similarities ($S_{pp}$). We also pass this batch to $G$ for encoding to generate another set of corresponding pairwise similarity matrices: $\sigma_{qp}$, $\sigma_{qn}$, $\sigma_{qq}$, $\sigma_{pp}$.


We use these similarities as indicators for removing potentially relevant examples to the evaluated query-positive pair ($qp^i$). That is, if any of the similarities in the similarity matrices $\sigma$, derived from vectors generated by $G$, is greater than the similarity $\sigma_{qp}^i$ of the query-positive pair, then we assume that these are examples that must not be considered as irrelevant. The remaining examples comprise $G_B$ and are then used as part of computing the contrastive loss, treating only examples with a similarity less than that of the query-positive pair as batch negatives.

This process is formalized as follows. Suppose at training timestep $t$ the model receives a batch of training data $B$ with $N_B$ samples, each comprising at least a pair of query-positive texts. For the $i^{th}$ sample in the batch, we compute the loss function using Equation \ref{eq:lg}. Where $q_i$ is the query text, $p_i^+$ is the positive text, and we define $G_B$ for this sample as samples having similarities less than $\sigma_{qp}^i$. This can be easily accomplished by masking the likely ``relevant" samples with $-\infty$, ensuring they will have no contribution to the loss.

\vspace{-0.2in}

\begin{equation*}
\begin{cases}
S_{qp}[\sigma_{qp} > \sigma_{qp}^i] = -\infty & \\

S_{qn}[\sigma_{qn} > \sigma_{qp}^i] = -\infty & \\

S_{qq}[\sigma_{qq} > \sigma_{qp}^i] = -\infty & \\

S_{pp}[\sigma_{pp} > \sigma_{qp}^i] = -\infty &
\end{cases}
\end{equation*}


We compute the loss using the similarities ($S$) from $M$ and the similarities ($\sigma$) from $G$ to select the negatives. It is also easy to see in this formulation that the selection of the in-batch negatives is independent of the originally assigned negative in the data. The number of negatives and the sources used in each training step also varies depending on the rate of potentially relevant examples the guide model identifies within the batch. This generalizes the bidirectional contrastive loss used by  \citeauthor{ni2021large, INSTRUCTOR} and relaxes the use of the full-batch as negatives \citeauthor{li2023towards} previously proposed. Notably, GISTEmbed shares some characteristics with the DCLR framework proposed by \citeauthor{zhou-etal-2022-debiased}, which adopts a complementary model. However, GISTEmbed offers greater robustness by eliminating the need to introduce new hyperparameters, which the DCLR requires, thereby enabling fully unsupervised and efficient fine-tuning without needing an expensive hyperparameter search. We show a diagram of GISTEmbed in Figure \ref{fig:gistembed}. In the rest of the paper, we interchangeably use GIST and GISTEmbed to refer to the framework.

\subsection{Advantages and limitations}

We see the proposed strategy as having two main value propositions. First, the framework does not require explicit negatives in the training data. It is sufficient only to have (query, positive) pairs in the training data, and we let the guide model mine the batch negatives for us. Second, the model can address potential data quality issues, especially when the training data has an incorrect assignment of positive and negative examples for a given query. In such a case, the guide model will ignore the assigned negative (supposed to be relevant) and pick other less relevant samples in the batch instead. This helps mitigate the potential confusion of the model.

The main limitation of this framework is its reliance on an existing model to guide during fine-tuning. However, recent strands of literature show the possibility of an agent or multiple LLM agents to self-improve \cite{huang-etal-2023-large, chan2023chateval}. It might, therefore, be worth exploring how two embedding models can be used to improve using GISTEmbed iteratively. Ultimately, the level of improvement will still be bottlenecked by the dataset available for training the model.



\section{Experimental setup}

To validate the effectiveness of GISTEmbed, we conduct a series of experiments on fine-tuning embedding models of various sizes. This section presents the methods and experimental details, including information on the datasets used. We also provide information on decisions adopted for the training strategy.

\subsection{Datasets}


\paragraph{MEDI dataset} We used the MEDI dataset introduced by \citeauthor{INSTRUCTOR} in their work on instruction fine-tuned text embeddings. The dataset contains a compilation of a large collection of corpora across many NLP tasks. It contains 1,450,000 triplets of query, positive, and negative examples mined unsupervisedly based on cosine similarities of the texts. Details to generate the triplets are discussed in the paper. Each triplet also comes with crafted instructions based on the task it belongs to. We only used the query, positive, and negative texts in this work and dropped the instructions.

\paragraph{MTEB classification dataset} The MTEB benchmark includes 12 classification datasets with training splits not used to assess the models' performance. As such, we considered these datasets as potentially beneficial sources of additional triplets to augment the MEDI dataset. Consequently, we augmented the MEDI dataset with the MTEB classification datasets we refer to as the MEDI+MTEBcls dataset. We selected 11 out of the 12 available datasets detailed in Appendix \ref{ax:mtebclass}. The Amazon Polarity dataset was not included due to its size. We mined triplets of samples from each classification dataset.

\paragraph{Mining for MTEB classification triplets} We apply the following algorithm for each dataset we included in our augmentation.

Our proposed algorithm for mining the triplets requires a meta-embedding model. We choose the \texttt{WhereIsAI/UAE-Large-V1} embedding model for selecting the triplets since it is the top-performing open embedding model, which has a reasonable size at the time of the work.

Triplets are mined such that the positive example for a query is selected from samples with the same label, while the negative example is chosen from the universe of samples that come from a different class from the query. The exact mining of triplets proceeds as follows:

First, assign one sample as the query. Then, we select the positive and negative samples to associate with the query with the algorithms below.

\textbf{Selecting the triplet positive}

\begin{itemize}
    \itemsep -0.3em
    \item Get the class the query belongs to and find all samples with the same class.
    \item Compute the cosine similarity, $\bm{\theta}$, of the query to these samples using the selected embedding model.
    \item Get the top-K$_p$ samples that have the highest similarity score.
    \item Compute a weighted probability over the top-K$_p$ samples based on their similarity scores to the query. We use a temperature parameter $\tau$ to control the distribution density.
            $$p_i = Softmax(topK(\bm{\theta}) / \tau)$$
    \item Assign a positive sample to the query by probabilistically selecting from the top-K$_p$ using the weighted probability.
\end{itemize}

\textbf{Selecting the triplet negative}

\begin{itemize}
    \itemsep -0.3em
    \item Get all samples having different classes as the query.
    \item Compute the similarity of the query to these samples and choose the top-K$_n$.
    \item Assign a uniform probability over the top-K$_n$.
    \item Probabilistically select from the top-K$_n$ based on the probability distribution and assign this as the negative sample forming the triplet.
\end{itemize}

In constructing the triplets, we set top-K$_p$ to 100 while we take top-K$_n$ to be all out-of-class samples. The temperature, $\tau$, was set to 0.05 for all datasets. We employed a uniform probability distribution for selecting the negatives to mitigate potential bias inherent in the reference model. This approach was critical because, although the meta embedding model used was specifically trained to enhance the similarity metrics among related concepts, it was not equally optimized to delineate the dissimilarity among unrelated concepts. By adopting a uniform distribution, we aim to ensure that the negative components of the triplets are not influenced by systematic representations of negative concepts learned by the reference model. Furthermore, systematically choosing the lowest-scored sample as the triplet negative may not be beneficial as this corresponds to an ``easy-negative", which can prevent the model from learning more robust representations.

\begin{table*}[ht]
\vspace{-0.1in}
\caption{Performance metrics (values in \%) across different models against the MTEB datasets. The table compares the metrics across task categories for the different base models, and the fine-tuned versions. We also present ablation runs for fine-tuning the models using only the data and standard contrastive training. The GISTEmbed provides universal improvements in the overall performance across tasks relative to the base models. The metrics show that smaller models benefit largely from being fine-tuned using GISTEmbed.}
\vskip 0.1in
\begin{center}
\begin{adjustbox}{max width=\textwidth}

    \begin{tabular}{lc|ccc|ccc|ccc}
        \toprule
        & & \multicolumn{3}{c}{all-MiniLM-L6-v2} & \multicolumn{3}{c}{bge-small-en-v1.5} & \multicolumn{3}{c}{bge-base-en-v1.5} \\
        & Metric & Base & Unguided & GIST.\footnotemark & Base & Unguided & GIST.\footnotemark & Base & Unguided & GIST.\footnotemark \\
        \midrule
        Classification (12)      & Acc. & 63.05 & 69.53 & \textbf{69.67} & 74.14 & \textbf{74.88} & 74.62 & 75.53 & \textbf{76.34} & 76.03 \\
        Clustering (11)           & V-meas. & \textbf{42.35} & 38.29 & 39.40 & 43.82 & 44.49 & \textbf{44.57} & 45.77 & \textbf{46.29} & 46.21 \\
        PairClassification (3)  & AP & 82.37 & 82.88 & \textbf{83.41} & 84.92 & 84.48 & \textbf{84.98} & \textbf{86.55} & 85.73 & 86.32 \\
        Reranking (4)            & MAP & \textbf{58.04} & 57.66 & 57.81 & 58.36 & 58.58 & \textbf{58.64} & 58.86 & 59.14 & \textbf{59.37} \\
        Retrieval (15)           & nDCG & 41.95 & 43.57 & \textbf{44.42} & \textbf{51.68} & 49.62 & 50.73 & \textbf{53.25} & 51.17 & 52.31 \\
        STS (10)                & Spear. & 78.90 & 79.89 & \textbf{80.34} & 81.59 & 82.63 & \textbf{83.19} & 82.40 & 82.76 & \textbf{83.51} \\
        Summarization (1)       & Spear. & 30.81 & \textbf{31.65} & 30.74 & 30.12 & \textbf{30.76} & 30.57 & \textbf{31.07} & 30.90 & 30.87 \\
        \midrule
        Total (56)               & Mean & 56.26 & 57.48 & \textbf{58.06} & 62.17 & 62.09 & \textbf{62.48} & 63.55 & 63.30 & \textbf{63.71} \\
        \bottomrule
    \end{tabular}
    \end{adjustbox}
\end{center}
\label{tab:performance_metrics}
\vskip -0.1in
\end{table*}

\footnotetext[2]{\href{https://huggingface.co/avsolatorio/GIST-all-MiniLM-L6-v2}{https://huggingface.co/avsolatorio/GIST-all-MiniLM-L6-v2}}

\footnotetext[3]{\href{https://huggingface.co/avsolatorio/GIST-small-Embedding-v0}{https://huggingface.co/avsolatorio/GIST-small-Embedding-v0}}

\footnotetext[4]{\href{https://huggingface.co/avsolatorio/GIST-Embedding-v0}{https://huggingface.co/avsolatorio/GIST-Embedding-v0}}

\begin{table}[ht]
\centering
\caption{Summary of model scores on the MTEB dataset for the bge-large-en-v1.5 model and GIST fine-tuned models (values in \%). We observe a large improvement in the performance for STS tasks, which aligns with the results from the other models. Similarly, the average score for retrieval tasks shows a significant decline. A closer look at the individual retrieval tasks points to the TRECCOVID dataset contributing largely to the decline, see Table \ref{tab:retrieval}}
\label{tab:performance_comparison}
\vskip 0.1in
\begin{adjustbox}{max width=0.48\textwidth}
\begin{tabular}{lc|cc}
\toprule
& & \multicolumn{2}{c}{bge-large-en-v1.5} \\
\textbf{Task} & \textbf{Metric} & \textbf{Base} & \textbf{GISTEmbed}\footnotemark \\
\midrule
Classification (12) & Acc. & 75.97 & \textbf{76.01} \\
Clustering (11) & V-meas. & 46.08 & \textbf{46.55} \\
Pair Classification (3) & AP & \textbf{87.12} & 86.70 \\
Reranking (4) & MAP & 60.03 & \textbf{60.05} \\
Retrieval (15) & nDCG & \textbf{54.29} & 53.44 \\
STS (10) & Spear. & 83.11 & \textbf{84.59} \\
Summarization (1) & Spear. & \textbf{31.61} & 30.96 \\
\midrule
Total (56) & Mean & 64.23 & \textbf{64.34} \\
\bottomrule
\end{tabular}
\end{adjustbox}
\vskip -0.1in
\end{table}

\footnotetext[5]{\href{https://huggingface.co/avsolatorio/GIST-large-Embedding-v0}{https://huggingface.co/avsolatorio/GIST-large-Embedding-v0}}

\subsection{Base models}

To measure the generalizability of the GISTEmbed framework, we perform fine-tuning of models with varying parameter numbers. We choose the following models, mainly based on the FlagEmbedding family of models \cite{xiao2023c}. These models are currently the top-performing open-sourced models with moderate-sized architectures based on the MTEB leaderboard. Within this family of models, the larger variant ranks second among the open models, with the guide model we employed being the top.

\paragraph{FlagEmbeddings} We selected the three models available comprising the FlagEmbedding models. The \texttt{bge-small-en-v1.5} model is a 33.4 million-parameter embedding model\footnote{\href{https://huggingface.co/BAAI/bge-small-en-v1.5}{https://huggingface.co/BAAI/bge-small-en-v1.5}}. The \texttt{bge-base-en-v1.5} model is a 109 million-parameter embedding model\footnote{\href{https://huggingface.co/BAAI/bge-base-en-v1.5}{https://huggingface.co/BAAI/bge-base-en-v1.5}}. And the \texttt{bge-large-en-v1.5} model is a 335 million-parameter embedding model\footnote{\href{https://huggingface.co/BAAI/bge-large-en-v1.5}{https://huggingface.co/BAAI/bge-large-en-v1.5}}. These models have been pre-trained and fine-tuned on a large collection of datasets. Using these models will allow us to get insights regarding the impact of GISTEmbed as the model size and base-performance scale.

\vspace{-0.15in}

\paragraph{Sentence transformers (all-MiniLM-L6-v2)} The
\texttt{all-MiniLM-L6-v2} model is a 22.7 million-parameter embedding model\footnote{\href{https://huggingface.co/sentence-transformers/all-MiniLM-L6-v2}{https://huggingface.co/sentence-transformers/all-MiniLM-L6-v2}}, which is part of the Sentence Transformers collection of embedding models \cite{reimers2019sentence}. We selected this model to study the potential of GISTEmbed in improving the performance of light-weight models. Unlike very large models, lightweight models are extremely useful for edge applications and systems with limited computing resources.

\subsection{Evaluation}

To evaluate the performance of the models, we use the Massive Text Embedding Benchmark (MTEB) dataset, which is a collection of NLP tasks that measures the performance of embedding models across general applications comprising 56 datasets \cite{muennighoff2022mteb}. The benchmark covers classification, clustering, pairwise classification, reranking, retrieval, semantic textual similarity (STS), and summarization tasks. A leaderboard hosted by HuggingFace also ranks the top-performing embedding models against the MTEB dataset\footnote{Leaderboard: \href{https://huggingface.co/spaces/mteb/leaderboard}{https://huggingface.co/spaces/mteb/leaderboard}}.

\begin{table}[ht!]
\centering
\caption{Comparison of performance metrics across different tasks for models fine-tuned using different datasets. The table indicates that the GISTEmbed fine-tuned models (guided) perform well compared to models fine-tuned without guidance. The models fine-tuned using the MEDI+MTEBcls dataset (+cls) also show improvements mainly in classification, pair classification, reranking, and summarization tasks. Using the MEDI dataset alone for fine-tuning proves to be marginally better for retrieval and STS tasks. Overall, leveraging GISTEmbed and using the MEDI+MTEBcls dataset provides a highly performant model.}
\label{tab:data-ablation}
\vskip 0.1in
    \begin{adjustbox}{max width=0.48\textwidth}
        \begin{tabular}{lcc|cc}
        \toprule
         & \multicolumn{2}{c}{Unguided} & \multicolumn{2}{c}{GISTEmbed} \\
        \cmidrule(lr){2-3} \cmidrule(lr){4-5}
        Dataset & MEDI & +cls & MEDI & +cls \\
        \midrule
        Classification (12) & 66.87 & 69.53 & 67.30 & \textbf{69.67} \\
        Clustering (11) & 38.29 & 38.29 & 39.29 & \textbf{39.40} \\
        PairClassification (3) & 82.41 & 82.88 & 83.12 & \textbf{83.41} \\
        Reranking (4) & 57.54 & 57.66 & 57.73 & \textbf{57.81} \\
        Retrieval (15) & 43.68 & 43.57 & \textbf{44.71} & 44.42 \\
        STS (10) & 80.04 & 79.89 & \textbf{80.47} & 80.34 \\
        Summarization (1) & 30.32 & \textbf{31.65} & 29.98 & 30.74 \\
        \midrule
        Total (56) & 56.91 & 57.48 & 57.60 & \textbf{58.06} \\
        \bottomrule
        \end{tabular}
    \end{adjustbox}
\vskip -0.1in
\end{table}

Evaluating the embedding models' performance across different task groups employs a diverse set of metrics, each tailored to the specific nature of the tasks involved. The accuracy metric is utilized for classification tasks, providing a straightforward measure of the model's ability to identify the category to which each instance belongs correctly. In clustering tasks, the validity of the embeddings is assessed using the V-measure \cite{rosenberg2007v}, a harmonic mean of precision and recall, which quantifies the effectiveness of the clustering by evaluating both the completeness and homogeneity of the clusters formed.

For tasks involving pairwise classification, precision is calculated based on the cosine similarities between pairs, offering insight into the model's capacity to identify relevant pairs within the dataset accurately. Reranking tasks, on the other hand, employ the Mean Average Precision (MAP), a metric that captures the model's ability to correctly order items in a way that higher relevance items appear before less relevant ones in the ranked list of search results.

Retrieval tasks utilize the Normalized Discounted Cumulative Gain (nDCG) at a cutoff of 10 (nDCG@10), which measures the model's effectiveness in retrieving highly relevant documents at the top of the ranking list, taking into account the position of each document in the result set. Finally, for Semantic Textual Similarity (STS) tasks and summarization, the Spearman correlation coefficient based on cosine similarity is used. This metric assesses the degree to which the model's similarity scores between texts align with human judgment, reflecting the model's proficiency in capturing the semantic relationships between pieces of text.

By adopting these specific metrics for each task group, the evaluation framework ensures a comprehensive and nuanced assessment of the embedding models' performance, highlighting their strengths and shortcomings in various language processing tasks.

\subsection{Training strategies and parameters}

In our model training process, we employed a learning rate of 5e-6, complemented by a warm-up ratio of 0.1 to adjust the learning rate at the beginning of training gradually. The optimization was carried out using the AdamW optimizer, with its beta parameters maintained at the default settings of (0.9, 0.999), and without applying weight decay. The training spanned over 100,000 steps, with each batch comprising 16 samples. For the contrastive loss, we set the temperature parameter, $\tau$, to 0.01, which is typical as used in \cite{INSTRUCTOR}. Additionally, all models were trained with a context length set to 512 tokens.

It is noteworthy that the Sentence Transformers model, as per its documentation\footnote{From the documentation: ``\href{https://huggingface.co/sentence-transformers/all-MiniLM-L6-v2\#intended-uses}{\textit{By default, input text longer than 256 word pieces is truncated.}}"}, imposes a default inference limit of 256 tokens for sequence length. However, our methodology deliberately extended this threshold to 512 tokens to ensure consistency with the context length specified during training. While the direct impact of this modification on the model's performance has not been empirically evaluated, this adjustment is strategically intended to exploit the entirety of the context made available during training, potentially enhancing the model's comprehension and output quality.

\section{Experiments}

\subsection{GISTEmbed comparison with unguided training}
We fine-tuned the base models on our training dataset using the GISTEmbed strategy and one using the standard method of fine-tuning. For training the models not using GISTEmbed, we adopted the improved contrastive loss using the full-batch as proposed in \cite{li2023towards}.

For reporting purposes, we identified and selected the checkpoints, specifically the evaluation steps, corresponding to the highest-scoring checkpoint among the unguided models. This selection process ensures that our reported outcomes likely represent the optimal performance achieved by the unguided models.

\begin{table*}[htbp]
\vspace{-0.1in}
    \centering
    \caption{Performance metrics of GIST fine-tuned all-MiniLM-L6-v2 across different training steps compared with the base model. All values are expressed as percentages. The trend shows that longer training improves the overall performance metrics for classification, retrieval, and STS tasks. There is no observed trend on the average performance scores for clustering, pair classification, reranking, and summarization tasks with increasing training length. The overall score shows increasing trend, mostly attributed to a large increase in performance of the embedding vectors generated by the model for classification tasks.}
    \vskip 0.1in
    \begin{center}
        \begin{tabular}{lc|cccccccc}
        \toprule
        & Base & \multicolumn{8}{c}{GISTEmbed} \\
        Training steps & & 15500 & 21000 & 40500 & 59500 & 75000 & 102000 & 171000 & 260000 \\
        \midrule
        Classification (12) & 63.05 & 64.56 & 65.22 & 67.19 & 68.32 & 68.66 & 69.67 & 71.17 & \textbf{72.72} \\
        Clustering (11) & \textbf{42.35} & 40.05 & 39.91 & 40.14 & 40.04 & 39.51 & 39.40 & 39.39 & 39.48 \\
        PairClassification (3) & 82.37 & 82.85 & 83.07 & \textbf{83.46} & 83.45 & 83.37 & 83.41 & 83.16 & 83.39 \\
        Reranking (4) & 58.04 & 58.00 & 58.08 & \textbf{58.10} & 57.77 & 57.72 & 57.81 & 57.96 & 57.94 \\
        Retrieval (15) & 41.96 & 43.19 & 43.72 & 44.21 & 44.12 & 44.05 & 44.42 & 45.00 & \textbf{45.12} \\
        STS (10) & 78.90 & 79.31 & 79.50 & 79.92 & 80.11 & 80.26 & 80.34 & 80.56 & \textbf{80.72} \\
        Summarization (1) & 30.81 & \textbf{31.82} & 31.45 & 31.02 & 30.56 & 30.63 & 30.74 & 30.65 & 31.22 \\
        \midrule
        Total (56) & 56.26 & 56.58 & 56.88 & 57.57 & 57.77 & 57.74 & 58.06 & 58.57 & \textbf{59.00} \\
        \bottomrule
        \end{tabular}
    \end{center}
\vskip -0.1in
\label{tab:trainingtrend}
\end{table*}

Our experiments show that using the GISTEmbed strategy improves the general performance of models in semantic similarity tasks, as Table \ref{tab:performance_metrics} shows. It is worth noting how the embeddings generated without using a guide model perform better in clustering for the bge-base model and classification tasks for both the bge-small and bge-base models. Most improvements are observed across tasks on smaller models---\texttt{all-MiniLM-L6-v2} and \texttt{bge-small-en-v1.5} models. While the performance on retrieval tasks, on average, looks abysmal, a view of the individual tasks provides a more nuanced perspective, Table \ref{tab:retrieval}. One of the more notable findings is the significant decrease in the nDCG@10 score for the TRECCOVID task across the fine-tuned models. This may be attributed to the lack of significant coverage of texts related to COVID in the training data used for fine-tuning. Ultimately, the strategy universally boosts the quality of embeddings relative to the respective base models used.


\subsection{MTEB classification triplets boost performance}

We assess the effect of augmenting the MEDI dataset with our mined triplets from the MTEB classification datasets by running ablations over the \texttt{all-MiniLM-L6-v2} model. The ablation experiments we performed tested fine-tuning the model using only the MEDI dataset. We also fine-tuned the base model using GISTEmbed and the MEDI dataset. We compare these runs against the fine-tuned models trained on the MEDI-MTEBcls dataset.

The results of these experiments, presented in Table \ref{tab:data-ablation}, demonstrate that the augmentation of the training data with the MTEBcls generally improves the model. Mainly, classification, pair classification, reranking, and summarization tasks show universal improvements. Models fine-tuned using only the MEDI data have better overall performance in retrieval and STS tasks.

Using the GISTEmbed fine-tuning and training on the MEDI+MTEBcls dataset shows significant improvement over the base model, with approximately a two percentage point increase in the overall score.

\begin{table*}[ht]
\vspace{-0.1in}
\centering
\caption{Task-level retrieval scores (nDCG@10) across different base and GISTEmbed fine-tuned models. Scores that are significantly better (absolute delta of 1\%) compared to the comparable model are shown in bold. Values are expressed as percentages. The (net top$_m$) column indicates the number of instances the GISTEmbed models have better retrieval scores minus the number of instances their respective base models outperformed them. Only scores that are significant are considered. The (\# top$_d$) row reports the number of datasets where a given model significantly outperforms the alternative. The table shows that the GISTEmbed-trained models suffer in the retrieval tasks as the base model used increases in size. It is interesting to see that some datasets that are part of MEDI didn't see improvement, especially the MSMARCO.}
\label{tab:retrieval}
\vskip 0.1in
    \begin{center}
        \begin{adjustbox}{max width=\textwidth}
            \begin{tabular}{lc|cc|cc|cc|cc}
            \toprule
            & & \multicolumn{2}{c}{all-MiniLM-L6-v2} & \multicolumn{2}{c}{bge-small-en-v1.5} & \multicolumn{2}{c}{bge-base-en-v1.5} & \multicolumn{2}{c}{bge-large-en-v1.5} \\
            \cmidrule(r){3-4} \cmidrule(l){5-6} \cmidrule(l){7-8} \cmidrule(l){9-10}
            & net top$_m$ & Base & GIST. & Base & GIST. & Base & GIST. & Base & GIST. \\
            \midrule
            ArguAna & $+2$ & 50.17 & \textbf{53.58} & 59.55 & \textbf{60.56} & 63.61 & 62.62 & 63.54 & 63.38 \\
            ClimateFEVER &  & 20.27 & \textbf{23.89} & 31.84 & 32.39 & 31.17 & 31.49 & \textbf{36.57} & 33.99 \\
            CQADupstackRetrieval+ & $+1$ & 41.32 & 41.26 & 39.05 & 39.88 & 42.35 & 43.20 & 42.23 & \textbf{43.44} \\
            DBPedia &  & 32.33 & \textbf{34.87} & 40.03 & 40.51 & 40.77 & 41.71 & \textbf{44.11} & 42.96 \\
            FEVER* & $+1$ & 51.93 & \textbf{70.80} & 86.64 & 87.27 & 86.29 & 86.65 & 87.18 & 86.55 \\
            FiQA2018 & $-1$ & 36.87 & 36.10 & \textbf{40.34} & 39.33 & 40.65 & 40.64 & 45.02 & 44.30 \\
            HotpotQA* & $-2$ & 46.51 & \textbf{51.63} & \textbf{69.94} & 66.94 & \textbf{72.60} & 68.92 & \textbf{74.10} & 70.46 \\
            MSMARCO* & $-1$ & 36.54 & 36.52 & 40.83 & 40.07 & 41.35 & 40.64 & \textbf{42.49} & 41.39 \\
            NFCorpus & $+1$ & 31.59 & 31.26 & 34.30 & \textbf{35.72} & 37.39 & 37.64 & 38.13 & 38.65 \\
            NQ* & $+1$ & 43.87 & \textbf{46.51} & \textbf{50.18} & 48.28 & 54.15 & 53.43 & 55.03 & \textbf{56.09} \\
            QuoraRetrieval &  & 87.56 & 88.03 & 88.78 & 88.56 & 88.90 & 88.81 & 89.07 & 88.98 \\
            SCIDOCS & $+3$ & 21.64 & 21.44 & 20.52 & \textbf{21.84} & 21.73 & \textbf{23.47} & 22.64 & \textbf{24.06} \\
            SciFact &  & \textbf{64.51} & 62.48 & 71.28 & 71.69 & 74.04 & \textbf{75.29} & 74.61 & 74.72 \\
            Touche2020 & $-2$ & 16.90 & \textbf{17.92} & \textbf{26.04} & 19.68 & \textbf{25.70} & 20.58 & \textbf{24.81} & 23.45 \\
            TRECCOVID & $-2$ & 47.25 & \textbf{50.05} & \textbf{75.90} & 68.28 & \textbf{78.07} & 69.61 & \textbf{74.82} & 69.13 \\
            \midrule
            Mean & & 41.95 & \textbf{44.42} & 51.68 & 50.73 & 53.25 & 52.31 & 54.29 & 53.44 \\
            \# top$_d$ & & 1 & 8 & 5 & 3 & 3 & 2 & 6 & 3 \\
            \bottomrule
            \end{tabular}
        \end{adjustbox}
    \end{center}
\vskip -0.1in
\end{table*}

\subsection{Effect of longer training}



We extended the training duration for the GISTEmbed-all-miniLM-L6-v2 model to examine the impact of increased data exposure on model performance. Annex \ref{ax:losscurve} details the resulting loss curve. Constraints on computational resources precluded similarly prolonging the training for other models. However, it would be beneficial for future studies to empirically investigate the effects of extended training durations on the performance of additional GISTEmbed models and models trained without the assistance of a guide model.

\begin{table*}[ht]
\vspace{-0.1in}
\centering
\caption{Comparison of model performance across tasks when augmenting the MEDI+MTEBcls dataset with triplets related to COVID search terms synthetically generated using GPT-4. The results show that the TRECCOVID task generally benefited from this augmentation. We also see marginal improvements in the overall performance of the models.
}
\label{tab:covq}
\vskip 0.1in
\begin{center}
\begin{adjustbox}{max width=\textwidth}
    \begin{tabular}{lcc|cc|cc}
    \toprule
    & \multicolumn{2}{c}{all-MiniLM-L6-v2} & \multicolumn{2}{c}{bge-small-en-v1.5} & \multicolumn{2}{c}{bge-base-en-v1.5} \\
    \cmidrule(r){2-3} \cmidrule(l){4-5} \cmidrule(l){6-7}
    Dataset & MEDI+MTEBcls & +COVq & MEDI+MTEBcls & +COVq & MEDI+MTEBcls & +COVq \\
    \midrule
    Classification (12) & 69.67 & 69.67 & 74.62 & 74.64 & 76.03 & 76.07 \\
    Clustering (11) & 39.40 & 39.09 & 44.57 & 44.69 & 46.21 & 46.44 \\
    PairClassification (3) & 83.41 & 83.50 & 84.98 & 84.97 & 86.32 & 86.35 \\
    Reranking (4) & 57.81 & 57.86 & 58.64 & 58.57 & 59.37 & 59.31 \\
    Retrieval (15) & 44.42 & 44.80 & 50.73 & 50.82 & 52.31 & 52.43 \\
    STS (10) & 80.34 & 80.36 & 83.19 & 83.13 & 83.51 & 83.55 \\
    Summarization (1) & 30.74 & 30.82 & 30.57 & 30.67 & 30.87 & 30.79 \\
    \midrule
    Total (56) & 58.06 & \textbf{58.11} & 62.48 & \textbf{62.51} & 63.71 & \textbf{63.80} \\
    \midrule
    TRECCOVID (nDCG@10) & 50.05 & \textbf{51.07} & 68.28 & \textbf{69.60} & 69.60 & \textbf{70.82} \\
    \bottomrule
    \end{tabular}
    \end{adjustbox}
    \end{center}
\vskip -0.1in
\end{table*}

The outcomes of this extended training are consolidated in Table \ref{tab:trainingtrend}, where we observed a general improvement in performance across classification, retrieval, and clustering tasks with prolonged training durations. Notably, classification tasks exhibited the most significant gains from extended training periods. This observation suggests that specific tasks, particularly those involving classification, may derive more significant benefits from longer training, potentially due to the increased opportunity for the model to refine the representations and optimize task-specific features. Furthermore, our previous findings indicate that incorporating triplets derived from the MTEB classification datasets into the MEDI dataset enhances classification scores. So, classification tasks may particularly benefit from extended training periods, as the model gains exposure to a larger volume of data relevant to these tasks.

\subsection{Task-specific augmentation improves the model}

As discussed earlier, a potential contributing factor to the observed performance degradation in the retrieval category, particularly in the TRECCOVID dataset, is the composition of the dataset used for fine-tuning. The diminished effectiveness of the fine-tuned models in this context may stem from an insufficient representation of COVID-related content within the fine-tuning dataset we used. Given that the base models were likely exposed to a broader dataset encompassing a range of contexts, including those relevant to COVID-19, their superior performance over the fine-tuned models hints at a crucial shortfall in fine-tuning---likelihood of model ``forgetfulness".

We investigated this hypothesis of model ``forgetfulness" by validating whether some of the lost performance can gained in the TRECCOVID task by leveraging augmented data related to COVID-19. We performed additional experiments leveraging 4,973 observations of synthetically generated triplets using GPT-4 based on search terms related to COVID-19 from Bing---we call this the COVq dataset\footnote{COVq raw triplets: \\\href{https://huggingface.co/datasets/avsolatorio/covid-bing-query-gpt4-avs_triplets}{https://huggingface.co/datasets/avsolatorio/covid-bing-query-gpt4-avs\_triplets}}. We use this synthetic data to augment the MEDI+MTEBcls dataset to compare whether improvements can be observed in the TRECCOVID task\footnote{MEDI+MTEBcls+COVq dataset:
\\\href{https://huggingface.co/datasets/avsolatorio/medi-data-mteb-covid-bing-query-gpt4-avs_triplets}{https://huggingface.co/datasets/avsolatorio/medi-data-mteb-covid-bing-query-gpt4-avs\_triplets}}. We detail the generation of the COVq dataset in Appendix \ref{ax:covq} and we present the results of this study in Table \ref{tab:covq}.

Our experiments show that the models fine-tuned with the dataset incorporating contents related to COVID-19 have universally improved performance in the TRECCOVID task, with at least a 1-percentage point lift in the nDCG@10 metric. However, the increase in scores in the TRECCOVID task is still lacking since the absolute scores are suboptimal compared to the original performance in the FlagEmbedding models. Nevertheless, the observed improvement in performance for the specific task we aimed to enhance upon incorporating task-related data emphasizes the intricate challenge posed by model fine-tuning: there's a risk of the model losing valuable information obtained from comprehensive upstream training datasets. This issue underscores the vital role of careful dataset selection during the fine-tuning stage, especially for applications with unique contextual requirements. Moreover, it necessitates a careful balance between boosting the model's specificity for specific tasks and maintaining the broad knowledge it has already acquired. Finally, despite being marginal, an enhancement in the overall performance across the models was also noted.

However, there are other potential reasons why the models we fine-tuned, whether employing GISTEmbed or not, have lower performance in certain retrieval tasks, not just in the TRECCOVID dataset. The HotpotQA dataset, for instance, is part of the MEDI dataset; however, the fine-tuned models still perform worse than the original models. One possible reason is the limited batch size we used for fine-tuning the models. Due to limited computing resources, we only trained the models for a batch size of 16, while we managed to use a batch size of 32 for the \texttt{bge-base-en-v1.5} model. It is, therefore, interesting to see how using a much larger batch size impacts the performance. The FlagEmbeddings have used batch sizes up to 19,200 in contrastive fine-tuning \cite{xiao2023c}. Having a larger batch size would yield more samples for contrast, which could provide the model nuance for improved relevance learning.

\subsection{GISTEmbed boosts Semantic Textual Similarity (STS) tasks}

Earlier, we show baseline comparison showing performance scores across various base and fine-tuned models, Table \ref{tab:performance_metrics}. Here, we focus on Semantic Textual Similarity (STS) tasks due to their significant impact on numerous downstream applications \cite{li20242d}.

We compare the performance of the fine-tuned model having 12 hidden layers against other models with comparable architecture, Table \ref{tab:sts}. The findings indicate a notable improvement in five of the seven STS tasks we used to test the models. This indicates that fine-tuning with the MEDI+MTEBcls dataset and employing GISTEmbed yields embeddings that result in better performance for tasks that demand semantic understanding.

\begin{table*}[ht]
\caption{Benchmarking of STS tasks performance across previous embedding models with comparable architectures (transformers). The results show that the GISTEmbed fine-tuned model (GIST-Embedding-v0) using the MEDI+MTEBcls dataset significantly outperforms other models in 5 out of 7 STS tasks. Consequently, this brings GIST-Embedding-v0 to have the highest average STS performance.}
\label{tab:sts}
\vspace{0.1in}
\centering
    \begin{adjustbox}{max width=\textwidth}
        \begin{tabular}{lccccccc|c}
        \hline
        Model & STS12 & STS13 & STS14 & STS15 & STS16 & STS-B & SICK-R & Avg. \\
        \hline
        GloVe \cite{reimers2019sentence} & 52.86 & 66.75 & 62.15 & 72.77 & 66.87 & 68.03 & 65.65 & 65.01 \\
        USE \cite{reimers2019sentence} & 64.49 & 67.80 & 64.61 & 76.83 & 73.18 & 74.92 & 76.69 & 71.22 \\
        SBERT \cite{reimers2019sentence} & 70.97 & 76.53 & 73.19 & 79.09 & 74.30 & 77.03 & 72.91 & 74.89 \\
        SimCSE \cite{gao-etal-2021-simcse} & 75.30 & 84.67 & 80.19 & 85.40 & 80.82 & 84.25 & 80.39 & 81.57 \\
        AnglE \cite{li2023angle} & 75.09 & 85.56 & 80.66 & 86.44 & 82.47 & 85.16 & 81.23 & 82.37 \\
        MRL (d = 768) \cite{li20242d} & 75.72 & 86.79 & 81.89 & 86.91 & 81.74 & 85.50 & 79.44 & 82.57 \\
        2DMSE (n = 12, d = 768)  \cite{li20242d} & 75.00 & 86.69 & 82.30 & 86.50 & 82.09 & 85.79 & 80.18 & 82.65 \\
        bge-base-en-v1.5 \cite{xiao2023c} & \textbf{78.03} & 84.19 & 82.27 & 87.96 & \textbf{85.48} & 86.42 & 80.30 & 83.52 \\
        \midrule
        GIST-Embedding-v0 (ours) & 76.12 & \textbf{87.85} & \textbf{83.39} & \textbf{89.43} & 85.35 & \textbf{87.32} & \textbf{81.29} & \textbf{84.39} \\
        \bottomrule
        
        \end{tabular}
    \end{adjustbox}
\end{table*}

\section{Discussion}

Our main findings point to smaller models having the largest benefit from using the GISTEmbed framework. Table \ref{tab:performance_metrics} underscores that smaller models, such as \texttt{all-MiniLM-L6-v2}, benefit significantly from GISTEmbed fine-tuning. This is particularly important for applications with limited computational resources, where deploying smaller yet efficient models is critical. The improvements in performance metrics suggest that GISTEmbed can make these smaller models competitive, offering a viable path to achieving high efficiency without compromising task performance.

Despite not using GISTEmbed, we still find performance improvements in fine-tuning existing models on readily available open datasets such as the MEDI+MTEBcls. However, the additional improvements achieved by using GISTEmbed provide evidence of shortcomings related to existing methods and potential issues in the training data that using a guide model was able to address dynamically. Consequently, researchers who have already trained embedding models using their proprietary datasets might find incorporating GISTEmbed into their workflows advantageous.

We see varied outcomes in the task-group level performance. While our results show, Table \ref{tab:performance_metrics}, that the average scores for tasks related to classification are higher when not using GISTEmbed for the \texttt{bge-small-en-v1.5} and \texttt{bge-base-en-v1.5} models, training the model further helps improve the performance on downstream machine learning tasks, as observed in Table \ref{tab:trainingtrend} for the sentence transformers model. Similarly, the retrieval task results reveal a complex picture, where the FlagEmbedding models outperform their fine-tuned counterparts, suggesting that the base models' state may be more aligned with task requirements for some of the retrieval tasks. For STS tasks, the guided versions of the models consistently achieve the highest scores, highlighting the benefit of targeted fine-tuning strategies in tasks requiring a nuanced understanding of text similarity.

Employing GISTEmbed demonstrates that models can gain significantly by utilizing a guide model to select in-batch negatives dynamically throughout the training process. However, the potential biases inherent in the guide model remain a challenge. Despite this, as more advanced and larger models emerge, GISTEmbed offers a strategic framework that allows smaller models to leverage these advancements, enhancing the quality of the embedding representations for improved performance in various downstream applications.

\section{Conclusion}

In this paper, we have presented GISTEmbed, a framework incorporating a guide model to dynamically select the in-batch negatives for the contrastive learning of embedding models to address potential data quality issues and noise in batch sampling. We mined training triplets from the MTEB classification corpora to augment the MEDI dataset; we call this the MEDI+MTEBcls dataset. Our experiments showed that using GISTEmbed and the MEDI+MTEBcls dataset generally improves the quality of embedding vectors across downstream applications measured using the MTEB benchmark. We also explored the implications of incorporating synthetically generated triplets using GPT-4 in the context of COVID-related search queries. Our analysis revealed a modest enhancement in retrieval performance; however, it notably fell short compared to the baseline retrieval scores achieved by the FlagEmbedding models utilized in the TRECCOVID task. We further demonstrate that GISTEmbed yields highly relevant embeddings for semantic textual similarity tasks benchmarked with other comparable models. Ultimately, adopting GISTEmbed has proven highly advantageous for smaller models, offering a promising pathway to strengthening artificial intelligence capabilities within resource-constrained environments.

\section*{Acknowledgements}

This work is supported by the "KCP IV - Exploring Data Use in the Development Economics Literature using Large Language Models (AI and LLMs)" project funded by the \href{https://www.worldbank.org/en/programs/knowledge-for-change}{Knowledge for Change Program (KCP)} of the World Bank - RA-P503405-RESE-TF0C3444. We also thank Olivier Dupriez for reviewing the manuscript.

The findings, interpretations, and conclusions expressed in this material are entirely those of the author(s). They do not necessarily represent the views of the International Bank for Reconstruction and Development/World Bank and its affiliated organizations, or those of the Executive Directors of the World Bank or the governments they represent.

\nocite{langley00}

\bibliography{gist_embed}
\bibliographystyle{icml2024}

\newpage
\appendix
\onecolumn

\section{MTEB Classification Datasets}

\begin{table}[H]
\label{ax:mtebclass}
\centering
\begin{tabular}{llcc}
\hline
\textbf{Dataset} & \textbf{Triplets (HuggingFace Datasets)} & \textbf{Classes} & \textbf{Number} \\ \hline
\href{https://huggingface.co/datasets/mteb/amazon_counterfactual}{Amazon Counterfactual} & \href{https://huggingface.co/datasets/avsolatorio/mteb-amazon_counterfactual-avs_triplets}{avsolatorio/mteb-amazon\_counterfactual-avs\_triplets} & 2 & 4.02k \\

\href{https://huggingface.co/datasets/mteb/amazon_massive_intent}{Amazon Massive Intent} & \href{https://huggingface.co/datasets/avsolatorio/mteb-amazon_massive_intent-avs_triplets}{avsolatorio/mteb-amazon\_massive\_intent-avs\_triplets} & 60 & 11.5k \\

\href{https://huggingface.co/datasets/mteb/amazon_massive_scenario}{Amazon Massive Scenario} & \href{https://huggingface.co/datasets/avsolatorio/mteb-amazon_massive_scenario-avs_triplets}{avsolatorio/mteb-amazon\_massive\_scenario-avs\_triplets} & 18 & 11.5k \\

\href{https://huggingface.co/datasets/mteb/amazon_reviews_multi}{Amazon Reviews} & \href{https://huggingface.co/datasets/avsolatorio/mteb-amazon_reviews_multi-avs_triplets}{avsolatorio/mteb-amazon\_reviews\_multi-avs\_triplets} & 5 & 200k \\

\href{https://huggingface.co/datasets/mteb/banking77}{Banking77} & \href{https://huggingface.co/datasets/avsolatorio/mteb-banking77-avs_triplets}{avsolatorio/mteb-banking77-avs\_triplets} & 77 & 10k \\

\href{https://huggingface.co/datasets/mteb/emotion}{Emotion} & \href{https://huggingface.co/datasets/avsolatorio/mteb-emotion-avs_triplets}{avsolatorio/mteb-emotion-avs\_triplets} & 6 & 16k \\

\href{https://huggingface.co/datasets/mteb/imdb}{IMDB} & \href{https://huggingface.co/datasets/avsolatorio/mteb-imdb-avs_triplets}{avsolatorio/mteb-imdb-avs\_triplets} & 2 & 25k \\

\href{https://huggingface.co/datasets/mteb/mtop_domain}{MTOP Domain} & \href{https://huggingface.co/datasets/avsolatorio/mteb-mtop_domain-avs_triplets}{avsolatorio/mteb-mtop\_domain-avs\_triplets} & 11 & 15.7k \\

\href{https://huggingface.co/datasets/mteb/mtop_intent}{MTOP Intent} & \href{https://huggingface.co/datasets/avsolatorio/mteb-mtop_intent-avs_triplets}{avsolatorio/mteb-mtop\_intent-avs\_triplets} & 113 & 15.7k \\

\href{https://huggingface.co/datasets/mteb/toxic_conversations_50k}{Toxic Conversations 50k} & \href{https://huggingface.co/datasets/avsolatorio/mteb-toxic_conversations_50k-avs_triplets}{avsolatorio/mteb-toxic\_conversations\_50k-avs\_triplets} & 2 & 50k \\

\href{https://huggingface.co/datasets/mteb/tweet_sentiment_extraction}{Tweet Sentiment Extraction} & \href{https://huggingface.co/datasets/avsolatorio/mteb-tweet_sentiment_extraction-avs_triplets}{avsolatorio/mteb-tweet\_sentiment\_extraction-avs\_triplets} & 3 & 27.5k \\
\hline
\end{tabular}
\caption{Overview of the MTEB classification datasets where additional training triplets were mined.}
\label{your_label_here}
\end{table}

\section{MEDI invalid triplet example}
\label{ax:invalidmedi}

\begin{minted}[
frame=lines,
framesep=2mm,
bgcolor=BrickRed!2,
baselinestretch=1.2,
fontsize=\scriptsize,
xleftmargin=10pt,
numbersep=5pt,
linenos
]{python}
# Example of a sample in the MEDI dataset having an irrelevant positive to the query.
{'query': [
    'Represent the scientific abstract for retrieving relevant citations;',
    'A sail-driven wind motor (SDWM) is described, in which a reciprocating load amenable to intermittent, \
    bidirectional drive (such as a water pump or air compressor) is driven by one or two arms, with each arm \
    driven by at least one conventional fore-and-aft rigged sail. Masts for the sails are mounted on beams \
    which are pivotally mounted on the arms. At least one mast and sail per beam, and at least one beam per arm \
    are required to drive the load. Mechanisms are described for controlling the sails to drive the load, and \
    to stop operation of the SDWM during excessive wind velocity. SDWMs for three types of sites are described: \
    land, shallow water, and deep water. Each arm is supported by at least one cross-arm, which rides on wheels \
    for a land site, or on at least one double-ended float for water sites. For all three types of sites, the \
    sail drag must be transferred to the ground, while the sail lift must be transferred to the load via the \
    arm. A mechanism is described for transferring the sail drag to the ground.'],
 'pos': [
    'Represent the scientific citation for retrieval;',
    'In this paper, we describe a multimodal application, called WiiNote, facilitating multi-user photo \
    annotation activity. The application allows up to 4 users to simultaneously annotating their pictures \
    adding either textual or vocal comments. Users use the Wii Remote device to select the whole picture or a \
    specific region of it to be annotated. Annotations can be either free or structured, i.e. based on a domain \
    specific data model expressed using MPEG7 standard or RDF language for ontology.'],
 'neg': [
    'Represent the scientific citation for retrieval;',
    'The major focus of the paper is the design and control of micro/picocellular systems supporting real-time \
    wireless connections subject to a guaranteed quality-of-service as defined by three metrics: call blocking, \
    call hand-off dropping, and forced call termination probability. The authors introduce the notion of the \
    cell-cluster and provide a model as well as an analytical methodology which can be used to design wireless \
    micro-cellular networks (in terms of \
    base station coverage), to allocate wireless spectrum (in terms of base station capacity), as well as to \
    perform wireless call admission control such that once a call is admitted to the system, it will enjoy a \
    predefined quality-of-service (in terms of call hand-off dropping and/or forced termination probability). \
    This wireless call control policy is simple enough that a high rate of mobile connection hand-offs can be \
    managed in a timely fashion.<<ETX>>'],
 'task_name': 'S2ORC_citations_abstracts'}
\end{minted}

\newpage

\section{Training loss curve}
\label{ax:losscurve}
\begin{figure*}[ht!]
    \vskip 0.2in
    \begin{center}
    \centerline{\includegraphics{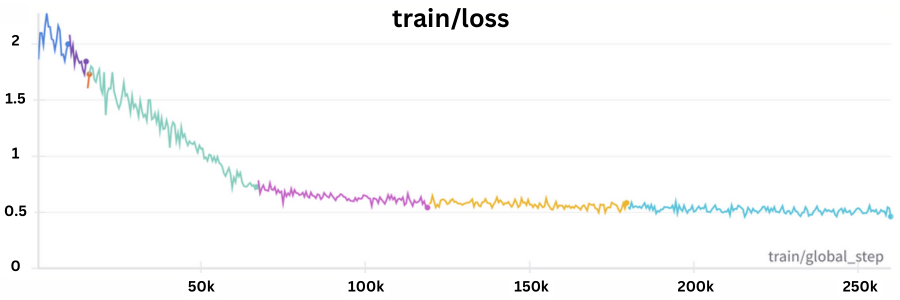}}
    \caption{Training loss curve for the all-MiniLM-L6-v2 model fine-tuning. The different colors simply show the continuation of training from checkpoints. There is a rapid phase of learning below the 50k steps mark. An inflection point at around 70k steps signal the beginning of saturation in the training.}
    \label{fig:all-mini-loss}
    \end{center}
    \vskip -0.2in
\end{figure*}

\section{Generating COVID-19 query triplets}
\label{ax:covq}

We use the \texttt{BingCoronavirusQuerySet} dataset to identify search terms related to COVID-19\footnote{BingCoronavirusQuerySet GitHub: \href{https://github.com/microsoft/BingCoronavirusQuerySet}{https://github.com/microsoft/BingCoronavirusQuerySet}}. First, we aggregate the search terms by country and limit only the terms to those tagged as explicitly related to COVID. Then, we sort the terms by descending popularity and get the top 100 for each country. We then combine the top terms by country to get the search terms we use as input to the LLM. The snippet below shows the parameters we use.

\begin{minted}[
frame=lines,
framesep=2mm,
bgcolor=BrickRed!2,
baselinestretch=1.2,
fontsize=\scriptsize,
xleftmargin=10pt,
numbersep=5pt,
linenos
]{python}
from openai import OpenAI
client = OpenAI()

response = client.chat.completions.create(
    model="gpt-4-turbo-preview",
    messages=[payload],
    temperature=1,
    max_tokens=4095,
    top_p=1,
    frequency_penalty=0,
    presence_penalty=0
)
\end{minted}

The payload contains the value \texttt{\{"role": "system", "content": \{content\}\}} where the content looks like the prompt shown below.

\begin{minted}[
frame=lines,
framesep=2mm,
bgcolor=BrickRed!2,
baselinestretch=1.2,
fontsize=\scriptsize,
xleftmargin=10pt,
numbersep=5pt,
linenos
]{md}
# System prompt
Generate a dataset of training triplets of COVID-19 queries, each with a positive example or relevant passage. \
The queries must be based on the search terms provided. Return the data as JSON records \
[{"s": <search terms>, "q": <query>, "p": <positive>}...]

search terms:
- term1
- term2
- ...
\end{minted}

The panel below shows the example synthetic data generated by the LLM.

\begin{minted}[
frame=lines,
framesep=2mm,
bgcolor=BrickRed!2,
baselinestretch=1.2,
fontsize=\scriptsize,
xleftmargin=10pt,
numbersep=5pt,
linenos
]{json}
[
    {
        "s": "coronavirus",
        "q": "What is coronavirus and how does it spread?",
        "p": "Coronavirus, also known as COVID-19, is a contagious virus primarily spread through droplets of \
        saliva or discharge from the nose when an infected person coughs or sneezes."
    },
    {
        "s": "coronavirus update",
        "q": "Can you provide the latest update on coronavirus situation worldwide?",
        "p": "As of the latest update, there are over 100 million confirmed cases of coronavirus globally, \
        with vaccinations underway in many countries to combat the pandemic."
    },
    {
        "s": "covid 19",
        "q": "What are the common symptoms of COVID-19?",
        "p": "Common symptoms of COVID-19 include fever, dry cough, and tiredness. Severe cases may experience \
        difficulty breathing or shortness of breath."
    }
]
\end{minted}




\section{Dataset versions}

Often, some improvements to datasets are introduced, so we list the specific versions of the dataset used in the experiments for replicability.

\begin{table}[ht]
\centering
\caption{Dataset Overview}
\begin{tabular}{l l l}
\hline
\textbf{Dataset Name} & \multicolumn{2}{l}{\textbf{Details}} \\
\hline
\multirow{2}{*}{MEDI dataset} & HF Dataset & \href{https://huggingface.co/datasets/avsolatorio/medi-data}{avsolatorio/medi-data} \\
 & Revision & 85c1250d939a02f277dfc4b33011bfd5a7f6dd07 \\
\hline
\multirow{2}{*}{MEDI+MTEBcls dataset} & HF Dataset & \href{https://huggingface.co/datasets/avsolatorio/medi-data-mteb\_avs\_triplets}{avsolatorio/medi-data-mteb\_avs\_triplets} \\
 & Revision & 238a0499b6e6b690cc64ea56fde8461daa8341bb \\
\hline
\multirow{2}{*}{MEDI+MTEBcls+COVIDq dataset} & HF Dataset & \href{https://huggingface.co/datasets/avsolatorio/medi-data-mteb-covid-bing-query-gpt4-avs\_triplets}{avsolatorio/medi-data-mteb-covid-bing-query-gpt4-avs\_triplets} \\
 & Revision & 7612b607f896cbf5d769dbe838ac83ce0807056b \\
\hline
\end{tabular}
\label{table:dataset_overview}
\end{table}

\end{document}